\documentclass[10pt,journal,compsoc]{IEEEtran}

\ifCLASSOPTIONcompsoc
  \usepackage[nocompress]{cite}
  \usepackage{graphicx}
\else
  \usepackage{cite}
\fi
\ifCLASSINFOpdf
\else
\fi

\usepackage{subfigure}
\usepackage{multirow}
\usepackage[square, comma, sort&compress, numbers]{natbib}

\begin{document}
%
\title{A3T-GCN: Attention Temporal Graph Convolutional Network for Traffic Forecasting}
%
%
%
%

\author{Jiawei Zhu, Yujiao Song, Lin Zhao and Haifeng Li*
\IEEEcompsocitemizethanks{\IEEEcompsocthanksitem H. Li, J. Zhu, Y. Song and L. Zhao are with School of Geosciences and Info-Physics, Central South University, Changsha 410083, China.}
}

\IEEEtitleabstractindextext{%
\begin{abstract}
Accurate real-time traffic forecasting is a core technological problem against the implementation of the intelligent transportation system. However,  it remains challenging considering the complex spatial and temporal dependencies among traffic flows. In the spatial dimension, due to the connectivity of the road network, the traffic flows between linked roads are closely related. In terms of the temporal factor, although there exists a tendency among adjacent time points in general, the importance of distant past points is not necessarily smaller than that of recent past points since traffic flows are also affected by external factors.  In this study, an attention temporal graph convolutional network (A3T-GCN) traffic forecasting method was proposed to simultaneously capture global temporal dynamics and spatial correlations. The A3T-GCN model learns the short-time trend in time series by using the gated recurrent units and learns the spatial dependence based on the topology of the road network through the graph convolutional network. Moreover, the attention mechanism was introduced to adjust the importance of different time points and assemble global temporal information to improve prediction accuracy. Experimental results in real-world datasets demonstrate the effectiveness and robustness of proposed A3T-GCN. The source code can be visited at https://github.com/lehaifeng/T-GCN/A3T.

\end{abstract}

}

\maketitle

\IEEEdisplaynontitleabstractindextext

%
\IEEEpeerreviewmaketitle

\ifCLASSOPTIONcompsoc
\IEEEraisesectionheading{\section{Introduction}\label{sec:introduction}}
\else
\section{Introduction}
\label{sec:introduction}
\fi

%
%
%
%
\IEEEPARstart{T}{raffic} forecasting is an important component of intelligent transportation systems and a vital part of transportation planning and management and traffic control \cite {Huang2005Dynamic,Jian2012Synthesis,Jing2004A,Gaoa2018Measuring}. Accurate real-time traffic forecasting has been a great challenge because of complex spatiotemporal dependencies. Temporal dependence means that traffic state changes with time, which is manifested by periodicity and tendency. Spatial dependence means that changes in traffic state are subject to the structural topology of road networks, which is manifested by the transmission of upstream traffic state to downstream sections and the retrospective effects of downstream traffic state on the upstream section\cite{Dong2012Spatial}. Hence, considering the complex temporal features and the topological characteristics of 
the road network is essential in realizing the traffic forecasting task.
\par Existing traffic forecasting models can be divided into parametric and non-parametric models. Common parametric models include historical average, time series \cite{Ahmed1979ANALYSIS,Hodge2014Short}, linear regression \cite{Sun2004Interval}, and Kalman filtering models\cite{Okutani1984Dynamic}. Although traditional parametric models use simple algorithms, they depend on stationary hypothesis. These models can neither reflect nonlinearity and uncertainty of traffic states nor overcome the interference of random events, such as traffic accidents. Non-parametric models can solve these problems well because they can learn the statistical laws of data automatically with adequate historical data. Common non-parametric models include k-nearest \cite{Altman1992An}, support vector regression (SVR) \cite{article,Fu2013Short}, fuzzy logic \cite{Yin2002Urban}, Bayesian network\cite{Sun2006A}, and neural network models.
\par Recently, deep neural network models have attracted wide attention from scholars because of the rapid development of deep learning \cite{Silver2017Mastering,Morav2017DeepStack}. Recurrent neural networks (RNNs), long short-term memory (LSTM) \cite{Graves1997Long}, and gated recurrent units (GRUs)\cite{Cho2014On} have been successfully utilized in traffic forecasting because they can use self-circulation mechanism and model temporal dependence \cite{Rui2016Using,Lint2002Freeway}. However, these models only consider the temporal variation of traffic state and neglect spatial dependence. Many scholars have introduced convolutional neural networks (CNNs) in their models to characterize spatial dependence remarkably. Wu et al. \cite{Wu2016Short} designed a feature fusion framework for short-term traffic flow forecasting by combining a CNN with LSTM. The framework captured the spatial characteristics of traffic flow through a one-dimensional CNN and explored short-term variations and periodicity of traffic flow with two LSTMs. Cao et al. \cite{Cao2017Interactive} proposed an end-to-end model called ITRCN, which transformed the interactive network flow to images and captured network flows using a CNN. ITRCN also extracted temporal features by using GRU. An experiment proved that the forecasting error of this method was 14.3\% and 13.0\% higher than those of GRU and CNN, respectively. Yu et al. \cite{Yu2017Spatiotemporal} captured spatial correlation and temporal dynamics by using DCNN and LSTM, respectively. They also proved the superiority of SRCN based on the investigation on the traffic network data in Beijing.
\par Although CNN is actually applicable to Euclidean data \cite{defferrard2016convolutional}, such as image and grids, it still has limitations in traffic networks, which possess non-Euclidean structures. In recent years, graph convolutional network (GCN) \cite{kipf2016semisupervised}, which can overcome the abovementioned limitations and capture structural characteristics of networks, has rapidly developed \cite{DBLP:journals/corr/LiYSL17, zhao2019t, yu2020forecasting}. In addition, RNNs and their variants use sequential processing over time and more apt to remember the latest information, thus are suitable to capture evolving short-term tendencies. While The importance of different time points cannot be distinguished only by the proximity of time. Mechanisms that are capable of learning global correlations are needed.
\par For this reason, an attention temporal GCN (A3T-GCN) was proposed for traffic forecasting task. The A3T-GCN combines GCNs and GRUs and introduces an attention mechanism\cite{xu2015attend, vaswani2017attention}. It not only can capture spatiotemporal dependencies but also ajust and assemble global variation information. The A3T-GCN is used for traffic forecasting on the basis of urban road networks.

\section{A3T-GCN}
\subsection{Definition of problems}
\par In this study, traffic forecasting is performed to predict future traffic state according to historical traffic states on urban roads. Generally, traffic state can refer to traffic flow, speed, and density. In this study, traffic state only refers to traffic speed.

Definition 1. Road network G: The topological structure of urban road network is described as $G=(V,E)$,where $V=\{v_1,v_2,\cdots,v_N\}$ is the set of road section, and N is the number of road sections.  $E$ is the set of edges, which reflects the connections between road sections. The whole connectivity information is stored in the adjacent matrix $A\in R^{N\times N}$, where rows and columns are indexed by road sections, and the value of each entry indicates the connectivity between corresponding road sections. The entry value is 0 if there is no existed link between roads and 1 (unweighted graph) or non-negative (weighted graph) if otherwise.

Definition 2. Feature matrix $X^{N\times P}$: Traffic speed on a road section is viewed as the attribute of network nodes, and it is expressed by the feature matrix $X\in R^{N\times P}$, where P is the number of node attribute features, that is, the length of historical time series. $X_{i}$ denotes the traffic speed in all sections at time i.

\par Therefore, the traffic forecasting modelling temporal and spatial dependencies can be viewed as learning a mapping function f on the basis of the road network G and feature matrix X of the road network. Traffic speeds of future T moments are calculated as follows:
\begin{equation}
\left[X_{t+1},\cdots,X_{t+T}\right] = f\left(G;\left(X_{t-n},\cdots,X_{t-1},X_{t}\right)\right) 
\end{equation}
where n is the length of a given historical time series, and T is the length of time series that needs to be forecasted.
\subsection{GCN model}
\par GCNs are semi-supervised models that can process graph structures. They are an advancement of CNNs in graph fields. GCNs have achieved many progresses in many applications, such as image classification \cite{Bruna2013Spectral}, document classification \cite{defferrard2016convolutional}, and unsupervised learning \cite{kipf2016semisupervised}. Convolutional mode in GCNs includes spectrum and spatial domain convolutions \cite{Bruna2013Spectral}. The former was applied in this study. Spectrum convolution can be defined as the product of signal x on the graph and figure filter $g_\theta(L)$,which is constructed in the Fourier domain:$g_\theta(L) \ast x = Ug_\theta(U^Tx)$, where $\theta$ is a model parameter, L is the graph Laplacian matrix, U is the eigenvector of normalized Laplacian matrix $L = I_N - D^{-\frac{1}{2}}AD^{-\frac{1}{2}} = U\lambda U^T$, and $U^Tx$ is the graph Fourier transformation of x. x can also be promoted to $X\in R^{N\times C}$, where C refers to the number of features.
\par Given the characteristic matrix X and adjacent matrix A, GCNs can replace the convolutional operation in anterior CNNs by performing the spectrum convolutional operation with consideration to the graph node and first-order adjacent domains of nodes to capture the spatial characteristics of graph. Moreover, hierarchical propagation rule is applied to superpose multiple networks. A multilayer GCN model can be expressed as:
\begin{equation}
    H^{\left(l+1\right)}=\sigma\left(\widetilde{D}^{-\frac{1}{2}}\,\widehat{A}\,\widetilde{D}^{-\frac{1}{2}}\,H^{\left(l\right)}\,\theta^{\left(l\right)}\right)
\end{equation}

\par where $\widetilde{A}=A+I_{N}$ is an adjacent matrix with self-connection structures, $I_{N}$ is an identity matrix, $\widetilde{D}$ is a degree matrix, $\widetilde{D}_{ii}=\sum_{j}\widetilde{A}_{ij}$, $H^{\left(l\right)}\in R^{N\times l}$ is the output of layer l, $\theta^{\left(l\right)}$ is the parameter of layer l, and $\sigma(\cdot)$ is an activation function used for nonlinear modeling.
\par Generally, a two-layer GCN model \cite{kipf2016semisupervised} can be expressed as:
\begin{equation}
f\left(X,A\right) = \sigma\left(\widehat{A}\,ReLU\left(\widehat{A}\, X\,W_{0}\right)W_{1}\right)
\end{equation}
\par where X is a feature matrix; A is the adjacent matrix; and $\widehat{A} = \widetilde{D}^{-\frac{1}{2}}\,\widetilde{A} \,\widetilde{D}^{-\frac{1}{2}}$is a preprocessing step, where $\widetilde{A}=A+I_{N}$ is the adjacent matrix of graph G with self-connection structure. $W_{0}\in R^{P\times H}$ is the weight matrix from the input layer to the hidden unit layer, where P is the length of time, and H is the number of hidden units. $W_{1}\in R^{H\times T}$ is the weight matrix from the hidden layer to the output layer. $f\left(X,A\right)\in R^{N\times T}$ denotes the output with a forecasting length of  $T$, and $ReLU()$is a common nonlinear activation function.

\par GCNs can encode the topological structures of road networks and the attributes of road sections simultaneously by determining the topological relationship between the central road section and the surrounding road sections. Spatial dependence can be captured on this basis. In a word, this study learned spatial dependence through the GCN model \cite{kipf2016semisupervised}.

\subsection{GRU model}
\par Temporal dependence of traffic state is another key problem that hinders traffic forecasting. RNNs are neural network models that process sequential data. However, limitations in long-term forecasting are observed in traditional RNNs because of disadvantages in gradient disappearance and explosion \cite{Bengio2002Learning}. LSTM \cite{Graves1997Long} and GRUs \cite{Cho2014On} are variants of RNNs that mediate the problems effectively. LSTM and GRUs basically have the same fundamental principles. Both models use gated mechanisms to maintain long-term information and perform similarly in various tasks \cite{chung2014empirical}. However, LSTM is more complicated, and it takes longer training time than GRUs, whereas GRU has a relatively simpler structure, fewer parameters, and faster training ability compared with LSTM.
\par In the present model, temporal dependence was captured by a GRU model. The calculation process is introduced as follows, where $h_{t-1}$ is the hidden state at t-1, $x_{t}$ is the traffic speed at the current moment, and $r_{t}$ is the reset gate to control the degree of neglecting the state information at the previous moment. Information unrelated with forecasting can be abandoned. If the reset gate outputs 0, then the traffic information at the previous moment is neglected. If the reset gate outputs 1, then the traffic information at the previous moment is brought into the next moment completely. $u_{t}$ is the update gate and is used to control the state information quantity at the previous moment that is brought into the current state. Meanwhile, $c_{t}$ is the memory content stored at the current moment, and $h_{t}$ is the output state at the current moment. 

\begin{equation}
u_{t}=\sigma(W_{u}\ast \left[X_{t},h_{t-1}\right]+b_{u})
\end{equation}
\begin{equation}
r_{t}=\sigma(W_{r}\ast \left[X_{t},h_{t-1}\right]+b_{r})
\end{equation}
\begin{equation}
c_{t}=\tanh(W_{c}\left[X_{t},(r_{t}*h_{t-1})\right]+b_{c})
\end{equation}
\begin{equation}
h_{t}=u_{t}*h_{t-1}+(1-u_{t})*c_{t}
\end{equation}

\par GRUs determine traffic state at the current moment by using hidden state at previous moment and traffic information at current moment as input. GRUs retain the variation trends of historical traffic information when capturing traffic information at current moment because of the gated mechanism. Hence, this model can capture dynamic temporal variation features from the traffic data, that is, this study has applied a GRU model to learn the temporal variation trends of the traffic state.

\subsection{Attention model}
\par Attention model is realized on the basis of encoder–decoder model. This model is initially used in neural machine translation tasks\cite{bahdanau2014neural}. Nowadays, attention models are widely applied in image caption generation \cite{xu2015attend}, recommendation system \cite{xiao2017attentional}, and document classification \cite{Pappas2017Multilingual}. With the rapid development of such models, existing attention models can be divided into multiple types, such as soft and hard attention\cite{bahdanau2014neural}, global and local attention\cite{Luong2015Effective}, and self-attention\cite{vaswani2017attention}. In the current study, a soft attention model was used to learn the importance of traffic information at every moment, and then a context vector that could express the global variation trends of traffic state was calculated for future traffic forecasting tasks.
\par Suppose that a time series $x_i(i=1,2,\cdots,n)$,where n is the time series length, is introduced. The design process of soft attention models is introduced as follows. First, the hidden states $h_i(i=1,2,\cdots,n)$ at different moments are calculated using CNNs (and their variants) or RNNs (and their variant), and they are expressed as $H=\{h_1,h_2,\cdots,h_n\}$.Second, a scoring function is designed to calculate the score/weight of each hidden state.  Third, an attention function is designed to calculate the context vector $(Ct)$ that can describe global traffic variation information. Finally, the final output results are obtained using the context vector. In the present study, these steps were followed in the design process, but a multilayer perception was applied as the scoring function instead.
\par Particularly, the characteristics $(h_i)$ at each moment were used as input when calculating the weight of each hidden state based on f. The corresponding outputs could be gained through two hidden layers. The weights of each characteristic $(\alpha_i)$ are calculated by a Softmax normalized index function (eq. (8)), where $w_{(1)}$ and $b_{(1)}$ are the weight and deviation of the first layer and $w_{(2)}$ and $b_{(2)}$ are the weight and deviation of the second layer, respectively.
\begin{equation}
e_{i} = w_{(2)}(w_{(1)}H + b_{(1)}) + b_{(2)}
\end{equation}

\begin{equation}
\alpha_{i} = \frac{\exp(e_{i})}{\sum_{k=1}^{n}\exp(e_{k})}
\end{equation}
\par Finally, the attention function was designed. The calculation process of the context vector $(C_t)$ that covers global traffic variation information is shown in Equation (10).
\begin{equation}
C_{t} =\sum_{i=1}^{n}\alpha_{i} \ast h_{i}
\end{equation}

\subsection{A3T-GCN model}
\par
\begin{figure*}[t]
	\centering
	\begin{center}
	   \includegraphics[width=1.0\linewidth]{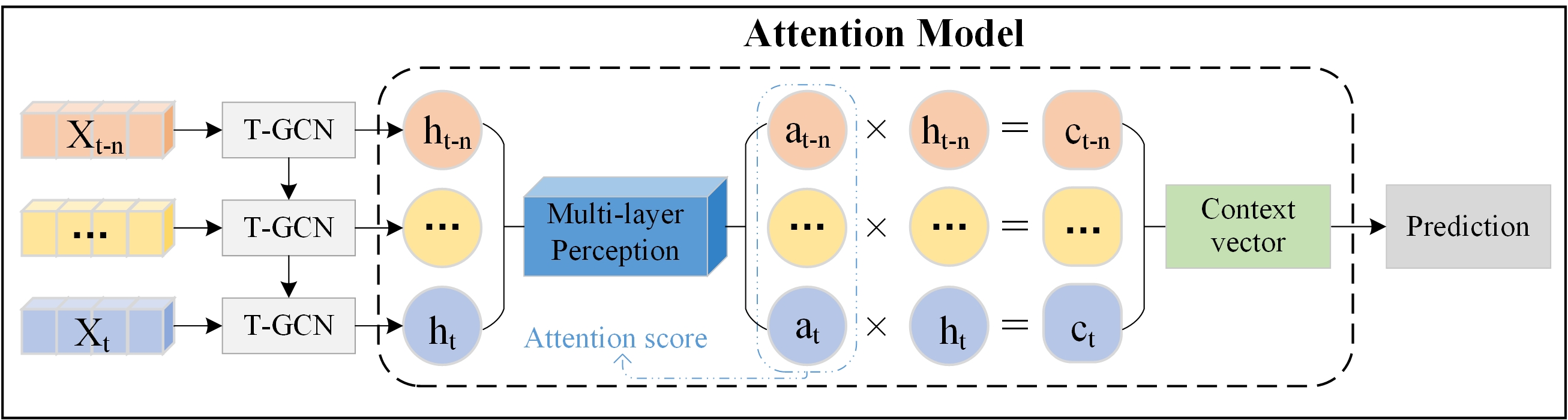}
	   \caption{A3T-GCN framework.}
	   \end{center}
	   \label{fig:1}
	\end{figure*}

 The A3t-GCN is a improvement of our previous work named T-GCN\cite{zhao2019t}. The attention mechanism was introduced to re-weight the influence of historical traffic states and thus to capture the global variation trends of traffic state. The model structure is shown in Fig. \ref{fig:1}.

\par A temporal GCN (T-GCN) model was constructed by combining GCN and GRU. n historical time series traffic data were inputted into the T-GCN model to obtain n hidden states (h) that covered spatiotemporal characteristics:$\{h_{t-n},\cdots,h_{t-1},h_t\}$.The calculation of the T-GCN is shown in eq. (11), where $h_{t-1}$ is the output at t-1. GC is the graph convolutional process. $u_t$ and $r_t$ are the update and reset gates at t, respectively. $c_t$ is the stored content at the current moment. $h_t$ is the output state at moment t, and W and b are the weight and the deviation in the training process, respectively.

\begin{equation}
u_{t} = \sigma(W_{u} \ast [GC(A,X_{t}),h_{t-1}]+b_{u})
\end{equation}

\begin{equation}
r_{t} = \sigma (W_{r} \ast [GC(A,X_{t}),h_{t-1}]+b_{r})
\end{equation}

\begin{equation}
c_{t} = \tanh (W_{c} \ast [GC(A,X_{t}),(r_{t} \ast h_{t-1})]+b_{c})
\end{equation}

\begin{equation}
h_{t} = u_{t} \ast h_{t-1}+(1-u_{t}) \ast c_{t})
\end{equation}
\par Then, the hidden states were inputted into the attention model to determine the context vector that covers the global traffic variation information. Particularly, the weight of each h was calculated by Softmax using a multilayer perception:$\{a_{t-n},\cdots,a_{t-1},a_t\}$.The context vector that covers global traffic variation information is calculated by the weighted sum. Finally, forecasting results were outputted using the fully connected layer. 
\par In sum, we proposed the A3T-GCN to realize traffic forecasting. The urban road network was constructed into a graph network, and the traffic state on different sections was described as node attributes. The topological characteristics of the road network were captured by a GCN to obtain spatial dependence. The dynamic variation of node attributes was captured by a GRU to obtain the local temporal tendency of traffic state. The global variation trend of the traffic state was then captured by the attention model, which was conducive in realizing accurate traffic forecasting.

\subsection{Loss function}
\par Training aims to minimize errors between real and predicted speed in the road network . Real and predicted speed on different sections at t are expressed by $Y$ and $\widehat{Y}$, respectively. Therefore, the objective function of A3T-GCN is shown as follows. The first term aims to minimize the error between real and predicted speed. The second term $L_{reg}$ is a normalization term, which is conducive to avoid overfitting. $\lambda$ is a hyper-parameter.
\begin{equation}
loss=\parallel Y_{t}-\widehat{Y_{t}}\parallel+\lambda L_{reg}
\end{equation}

\section{Experiments}

\subsection{Data Description}
\par Two real-world traffic datasets, namely, taxi trajectory dataset (SZ\_taxi) in Shenzhen City and loop detector dataset (Los\_loop) in Los Angeles, were used. Both datasets are related with traffic speed. Hence, traffic speed is viewed as the traffic information in the experiments. SZ\_taxi dataset is the taxi trajectory of Shenzhen from Jan. 1 to Jan. 31, 2015. In the present study, 156 major roads of Luohu District were selected as the study area. Los\_loop dataset is collected in the highway of Los Angeles County in real time by loop detectors. A total of 207 sensors along with their traffic speed from Mar. 1 to Mar. 7, 2012 were selected.
\subsection{Evaluation Metrics}
\par To evaluate the prediction performance of the model, the error between real traffic speed and predicted results is evaluated on the basis of the following metrics:
\par (1) Root Mean Squared Error (RMSE):
\begin{equation}
RMSE=\sqrt{\frac{1}{M N}\sum_{j=1}^{M}\sum_{i=1}^{N}(y_{i}^{j}-\widehat{y_{i}^{j}})^{2}}
\end{equation}

\par (2) Mean Absolute Error (MAE):
\begin{equation}
MAE=\frac{1}{M N}\sum_{j=1}^{M}\sum_{i=1}^{N}\left|y_{i}^{j}-\widehat{y_{i}^{j}}\right|
\end{equation}

\par (3) Accuracy:
\begin{equation}
Accuracy=1-\frac{\parallel Y-\widehat{Y}\parallel_{F}}{\parallel Y\parallel_{F}}
\end{equation}

\par (4) Coefficient of Determination ($R^{2}$):
\begin{equation}
R^{2}=1-\frac{\sum_{j=1}^{M}\sum_{i=1}^{N}(y_{i}^{j}-\widehat{y_{i}^{j}})^{2}}{\sum_{j=1}^{M}\sum_{i=1}^{N}(y_{i}^{j}-\bar{Y})^{2}}
\end{equation}

(5) Explained Variance Score ($var$):
\begin{equation}
var=1-\frac{Var\left\{Y-\widehat{Y}\right\}}{Var\left\{Y\right\}}
\end{equation}

where $y_{i}^{j}$ and $\widehat{y_{i}^{j}}$ are the real and predicted traffic information of temporal sample j on road i, respectively. N is the number of nodes on road. M is the number of temporal samples. $Y$ and $\widehat{Y}$ are the set of $y_{i}^{j}$ and $\widehat{y_{i}^{j}}$ respectively, and $\bar{Y}$ is the mean of $Y$.

Particularly, RMSE and MAE are used to measure prediction error. Small RMSE and MASE values reflect high prediction precision. Accuracy is used to measure forecasting precision, and high accuracy value is preferred. $R^{2}$ and $var$ calculate the correlation coefficient, which measures the ability of the prediction result to represent the actual data: the larger the value is, the better the prediction effect is.

\subsection{ Experimental result analysis}

\par The hyper-parameters of A3T-GCN include learning rate, epoch, and number of hidden units. In the experiment, learning rate and epoch were manually set on the basis of experiences as 0.001 and 5000 for both datasets. As for the number of hidden units, we set it to 64 and 100 for SZ\_taxi and Los\_loop, respectively.
\begin{table*}
	\caption{The prediction results of the T-GCN model and other baseline methods on SZ-taxi and Los-loop datasets.}
	\centering
	\resizebox{160mm}{40mm}{
	\renewcommand{\arraystretch}{1.3}
	\begin{tabular}{c|c|cccccc|cccccc}
		\hline
		\multirow{2}{*}{T}&
		\multirow{2}{*}{Metric}&
		\multicolumn{6}{c|}{SZ-taxi}&
		\multicolumn{6}{c}{Los-loop} \\
		\cline{3-14}
		&&HA&ARIMA&SVR&GCN&GRU&AT-GCN&HA&ARIMA&SVR&GCN&GRU&AT-GCN\\
		\hline\hline
		\multirow{5}*{15min}
		&$RMSE$&4.2951&7.2406&4.1455&5.6596&3.9994& \textbf{3.8989}&7.4427&10.0439&6.0084&7.7922&5.2182&\textbf{5.0904}\\
		&$MA$E&2.7815&4.9824&2.6233&4.2367&\textbf{2.5955}&2.6840&4.0145&7.6832&3.7285&5.3525&\textbf{3.0602}&3.1365\\
		&$Accuracy$&0.7008&0.4463&0.7112&0.6107&0.7249&\textbf{0.7318}&0.8733&0.8275&	0.8977&0.8673&0.9109&\textbf{0.9133}\\
		&$R^{2}$&0.8307&$\ast$&0.8423&0.6654&0.8329&\textbf{0.8512}&0.7121&0.0025&0.8123&0.6843&0.8576&\textbf{0.8653}\\
		&$var$&0.8307&0.0035&0.8424&0.6655&0.8329&\textbf{0.8512}&0.7121&$\ast$&0.8146&0.6844&0.8577&\textbf{0.8653}\\
		\hline
		\multirow{5}*{30min}
		&$RMS$E&4.2951&6.7899&4.1628&5.6918&4.0942&\textbf{3.9228}&7.4427&9.3450&6.9588&8.3353&6.2802&\textbf{5.9974}\\
		&$MAE$&2.7815&4.6765&\textbf{2.6875}&4.2647&2.6906&2.7038&4.0145&7.6891&3.7248&5.6118&\textbf{3.6505}&3.6610\\
		&$Accuracy$&0.7008&0.3845&0.7100&0.6085&0.7184&\textbf{0.7302}&0.8733&0.8275&0.8815&0.8581&0.8931&\textbf{0.8979}\\
		&$R^{2}$&0.8307&$\ast$&0.8410&0.6616&0.8249&\textbf{0.8493}&0.7121&0.0031&0.7492&0.6402&0.7957&\textbf{0.8137}\\
		&$var$&0.8307&0.0081&0.8413&0.6617&0.8250&\textbf{0.8493}&0.7121&$\ast$&0.7523&0.6404&0.7958&\textbf{0.8137}\\
		\hline
		\multirow{5}*{45min}
		&$RMSE$&4.2951&6.7852&4.1885&5.7142&4.1534&\textbf{3.9461}&7.4427&10.0508&7.7504&8.8036&7.0343&\textbf{6.6840}\\
		&$MAE$&2.7815&4.6734&\textbf{2.7359}&4.2844&2.7743&2.7261&4.0145&7.6924&4.1288&5.9534&\textbf{4.0915}&4.1712\\
		&$Accuracy$&0.7008&0.3847&0.7082&0.6069&0.7143&\textbf{0.7286}&0.8733&0.8273&0.8680&0.8500&0.8801&\textbf{0.8861}\\
		&$R^{2}$&0.8307&$\ast$&0.8391&0.6589&0.8198&\textbf{0.8474}&0.7121&$\ast$&0.6899&0.5999&0.7446&\textbf{0.7694}\\
		&$var$&0.8307&0.0087&0.8397&0.6590&0.8199&\textbf{0.8474}&0.7121&0.0035&0.6947&0.6001&0.7451&\textbf{0.7705}\\	
		\hline
		\multirow{5}*{60min}
		&$RMSE$&4.2951&6.7708&4.2156&5.7361&4.0747&\textbf{3.9707}&7.4427&10.0538&8.4388&9.2657&7.6621&\textbf{7.0990}\\
		&$MAE$&2.7815&4.6655&2.7751&4.3034&\textbf{2.7712}&2.7391&4.0145&7.6952&\textbf{4.5036}&6.2892&4.5186&4.2343\\
		&$Accuracy$&0.7008&0.3851&0.7063&0.6054&0.7197&\textbf{0.7269}&0.8733&0.8273&0.8562&0.8421&0.8694&\textbf{0.8790}\\
		&$R^{2}$&0.8307&$\ast$&0.8370&0.6564&0.8266&\textbf{0.8454}&0.7121&$\ast$&0.6336&0.5583&0.6980&\textbf{0.7407}\\
		&$var$&0.8307&0.0111&0.8379&0.6564&0.8267&\textbf{0.8454}&0.7121&0.0036&0.5593&0.5593&0.6984&\textbf{0.7415}\\
		\hline
	\end{tabular}}
	\label{table}
\end{table*}

\par In the present study, 80\% of the traffic data are used as the training set, and the remaining 20\% of the data are used as the test set. The traffic information in the next 15, 30, 45, and 60 min is predicted. The predicted results are compared with results from the historical average model (HA), auto-regressive integrated moving average model (ARIMA), SVR, GCN model, and GRU model. The A3T-GCN is analyzed from perspectives of precision, spatiotemporal prediction capabilities, long-term prediction capability, and global feature capturing capability.
\par (1) High prediction precision. Table \ref{table} shows the comparisons of different models and two real datasets in terms of the prediction precision of various traffic speed lengths. The prediction precision of neural network models (e.g., A3T-GCN and GRU) is higher than those of other models (e.g., HA, ARIMA, and SVR). With respect to 15-minute time series, the RMSE and accuracy of HA are approximately 9.22\% higher and 4.24\% lower than those of A3T-GCN, respectively. The RMSE and accuracy of ARIMA are approximately 46.15\% higher and 39.01\% lower than those of A3T-GCN, respectively. The RMSE and accuracy of SVR are  approximately 5.95\% higher and 2.81\% lower than those of A3T-GCN, respectively. Compared with GRU, The RMSE and accuracy of HA is approximately 6.88\% higher and 3.32\% lower than those of GRU, respectively. The RMSE and accuracy of ARIMA are approximately 44.76\%  and 38.07\%, respectively. The RMSE and accuracy of SVAR are approximately 3.52\% and 1.87\%, respectively. These results are mainly caused by the poor nonlinear fitting abilities of HA, ARIMA, and SVAR to complicated changing traffic data. Processing long-term non-stationary data is difficult when ARIMA is used. Moreover, ARIMA is gained by averaging the errors of different sections. The data of some sections might greatly fluctuate to increase the final error. Hence, ARIMA shows the lowest forecasting accuracy.
\par Similar conclusions could be drawn for Los\_loop. In a word, A3T-GCN model can obtain the optimal prediction performance of all metrics in two real datasets, thereby proving the validity and superiority of A3T-GCN model in spatiotemporal traffic forecasting tasks. 

(2) Effectiveness of modelling both spatial and temporal dependencies. To test the benefits brought by depicting the spatiotemporal characteristics of traffic data simultaneously in A3T-GCN, the model is compared with GCN and GRU.

\begin{figure}[t]
\begin{center}
  \includegraphics[width=1.0\linewidth]{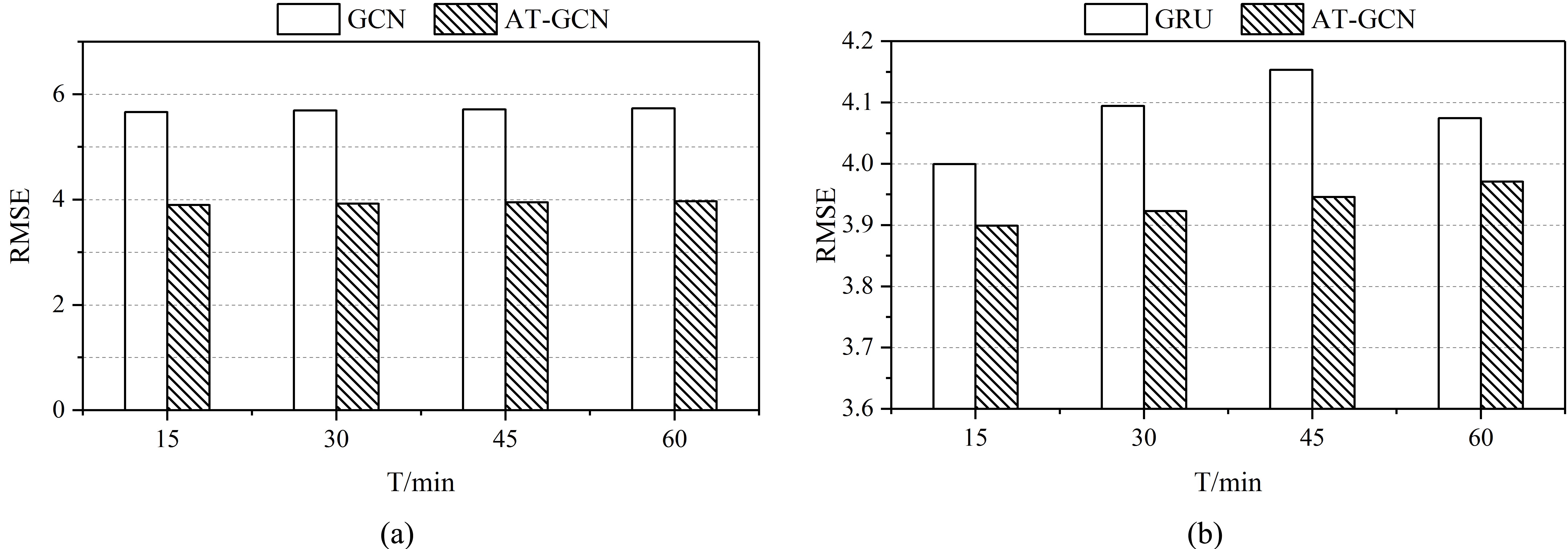}
  \end{center}
  \caption{ SZ-taxi: Spatiotemporal prediction capabilities.}
\label{fig:4}
\end{figure}

\begin{figure}[t]
\begin{center}
  \includegraphics[width=1.0\linewidth]{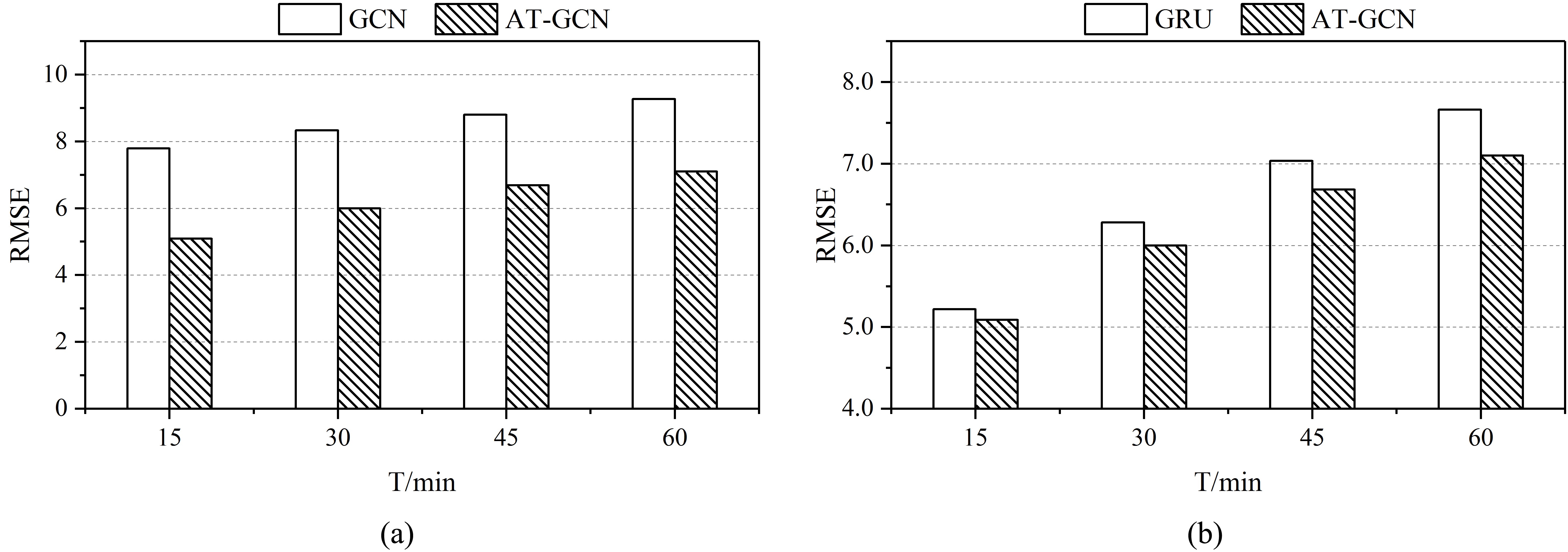}
  \end{center}
  \caption{Los-loop: Spatiotemporal prediction capabilities.}
\label{fig:5}
\end{figure}

\par Fig. \ref{fig:4} shows the results based on SZ\_taxi. Compared with GCN (considering spatial characteristics only), A3T-GCN achieves approximately 31.11\%, 31.08\%, 30.94\%, and 30.78\% lower RMSEs in 15, 30, 45, and 60 minutes of traffic forecasting time series, respectively. In sum, the prediction error of A3T-GCN is kept lower than that of GCN in 15, 30, 45, and 60 minutes of traffic forecasting. Therefore, the A3T-GCN can capture spatial characteristics.
\par Compared with GRU (considering temporal characteristics only), A3T-GCN achieves approximately 2.51\% lower RMSE in 15 minutes traffic forecasting, approximately 4.19\% lower RMSE in 30 minutes traffic forecasting, approximately 4.99\% lower RMSE in 45 minutes time series, and approximately 2.55\% lower RMSE in 60 minutes time series. In sum, the prediction error of A3T-GCN is kept lower than that of GRU in 15, 30, 45, and 60 minutes traffic forecasting. Therefore, the A3T-GCN can capture temporal dependence.
\par Results based on Los\_loop, which are similar with those based on SZ\_taxi, are shown in Fig. \ref{fig:5}. In short, the A3T-GCN has good spatiotemporal prediction capabilities. In other words, A3T-GCN model can capture the spatial topological characteristics of urban road networks and the temporal variation characteristics of traffic state simultaneously.

\par
\begin{figure}
\begin{center}
   \includegraphics[width=1.0\linewidth]{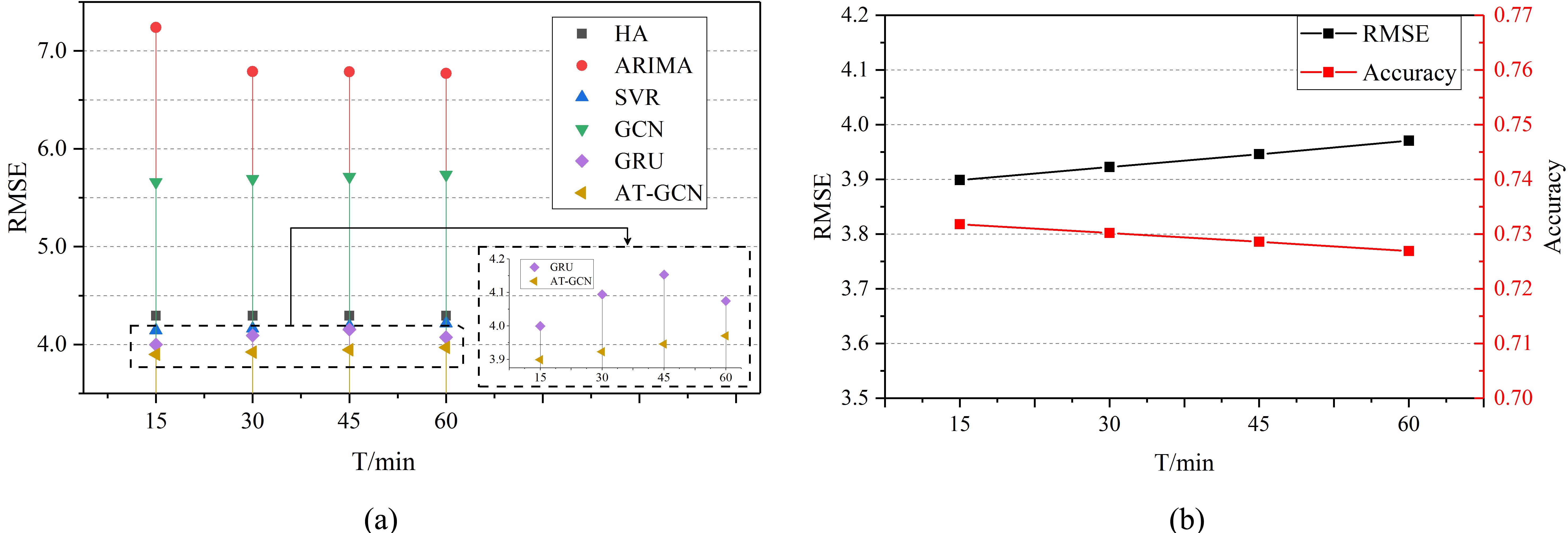}
   \end{center}
   \caption{ SZ-taxi: Long-term prediction capability.}
\label{fig:6}
\end{figure}

 (3) Long-term prediction capability. Long-term prediction capability of A3T-GCN was tested by traffic speed forecasting in 15, 30, 45, and 60 minutes prediction horizon.
Forecasting results based on SZ-taxi are shown in Fig. \ref{fig:6}. The RMSE comparison of different models under different lengths of time series is shown in Fig. \ref{fig:6}(a). The RMSE of the A3T-GCN is the lowest under all lengths of time series. The variation trends of RMSE and accuracy, which reflects prediction error and precision, respectively, of the A3T-GCN under different lengths of time series are shown in Fig. \ref{fig:6}(b). RMSE increases as the length of time series increases, whereas accuracy declines slightly and shows certain stationary.
\par
\begin{figure}
\begin{center}
   \includegraphics[width=1.0\linewidth]{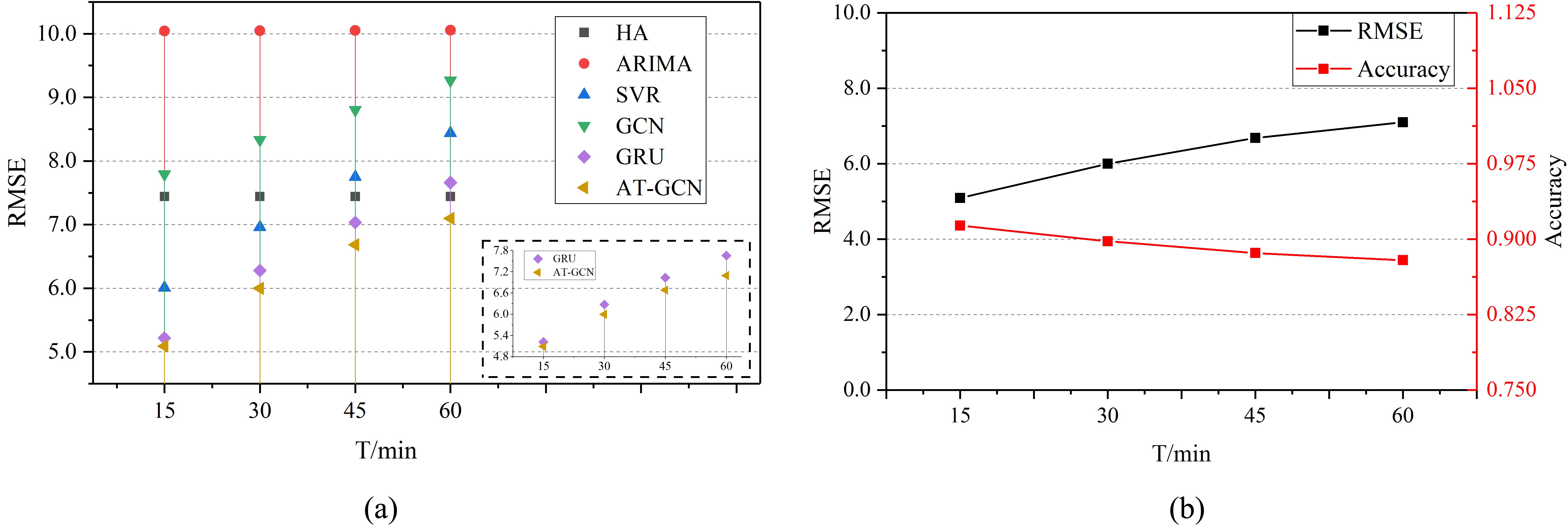}
   \end{center}
   \caption{Los-loop: Long-term prediction capability.}
\label{fig:7}
\end{figure}

 The forecasting results based on Los\_loop are shown in Fig. \ref{fig:7}, and consistent laws are found. In sum, A3T-GCN has good long-term prediction capability. It can obtain high accuracy by training for 15, 30, 45, and 60 minutes prediction horizon. Forecasting results of A3T-GCN change slightly with changes in length of time series, thereby showing certain stationary. Therefore, the A3T-GCN is applicable to short-term and long-term traffic forecasting tasks.

\begin{table}
\footnotesize
	\caption{Comparison of forecasting results between A3T-GCN and T-GCN under different lengths of time series based on SZ-taxi and Los-loop.}
	\centering
	\renewcommand{\arraystretch}{1}
	
	\begin{tabular}{c|c|cc|cc}
		\hline
		\multirow{2}{*}{T}&
		\multirow{2}{*}{Metric}&
		\multicolumn{2}{c|}{SZ-taxi}&
		\multicolumn{2}{c}{Los-loop} \\
		\cline{3-6}
		&&T-GCN&AT-GCN&T-GCN&AT-GCN\\

		\hline\hline
		\multirow{5}*{15min}
		&$RMSE$&3.9325&\textbf{3.8989}&5.1264&\textbf{5.0904}\\
		&$MAE$&2.7145&\textbf{2.6840}&3.1802&\textbf{3.1365}\\
		&$Accuracy$&0.7295&\textbf{0.7318}&0.9127&\textbf{0.9133}\\
		&$R^{2}$&0.8539&\textbf{0.8512}&0.8634&\textbf{0.8653}\\
		&$var$&0.8539&\textbf{0.8512}&0.8634&\textbf{0.8653}\\
		\hline
		\multirow{5}*{30min}
		&$RMS$E&3.9740&\textbf{3.9228}&6.0598&\textbf{5.9974}\\
		&$MAE$&2.7522&\textbf{2.7038}&3.7466&\textbf{3.6610}\\
		&$Accuracy$&0.7267&\textbf{0.7302}&0.8968&\textbf{0.8979}\\
		&$R^{2}$&0.8451&\textbf{0.8493}&0.8098&\textbf{0.8137}\\
		&$var$&0.8451&\textbf{0.8493}&0.8100&\textbf{0.8137}\\
		\hline
		\multirow{5}*{45min}
		&$RMSE$&3.9910&\textbf{3.9461}&6.7065&\textbf{6.684}\\
		&$MAE$&2.7645&\textbf{2.7261}&4.1158&\textbf{4.1712}\\
		&$Accuracy$&0.7255&\textbf{0.7286}&0.8857&\textbf{0.8861}\\
		&$R^{2}$&0.8436&\textbf{0.8474}&0.7679&\textbf{0.7694}\\
		&$var$&0.8436&\textbf{0.8474}&0.7684&\textbf{0.7705}\\	
		\hline
		\multirow{5}*{60min}
		&$RMSE$&4.0099&\textbf{3.9707}&7.2677&\textbf{7.099}\\
		&$MAE$&2.7860&\textbf{2.7391}&4.6021&\textbf{4.2343}\\
		&$Accuracy$&0.7242&\textbf{0.7269}&0.8762&\textbf{0.8790}\\
		&$R^{2}$&0.8421&\textbf{0.8454}&0.7283&\textbf{0.7407}\\
		&$var$&0.8421&\textbf{0.8454}&0.7290&\textbf{0.7415}\\
		\hline
	\end{tabular}
	\label{table2}
\end{table}
\par (4) Effectiveness of introducing attention to capture global variation. A3T-GCN and T-GCN were compared to test the superiority of capturing global variation. Results are shown in Table \ref{table2}. A3T-GCN model shows approximately 0.86\% lower RMSE and approximately 0.32\% higher accuracy than T-GCN model under 15 minutes time series, approximately 1.31\% lower RMSE and approximately 0.48\% higher accuracy under 30 minutes time series, approximately 1.14\% lower RMSE and approximately 0.43\% higher accuracy under 45 minutes traffic forecasting, and approximately 0.99\% lower RMSE and approximately 0.37\% higher accuracy under 60 minutes time series.
\par Hence, the prediction error of A3T-GCN is lower than that of T-GCN, but the accuracy of the former is higher under different horizons of traffic forecasting, thereby proving the global feature capturing capability of the A3T-GCN model.

\subsection{ Perturbation analysis}
\par Noise is inevitable in real-world datasets. Therefore, perturbation analysis is conducted to test the robustness of A3T-GCN. In this experiment, two types of random noises are added to the traffic data. Random noise obeys Gaussian distribution $N\in (0,\sigma^2)$, where $\sigma\in(0.2,0.4,0.8,1,2)$, and Poisson distribution $P(\lambda)$ where $\lambda\in(1,2,4,8,16)$. The noise matrix values are normalized to [0,1].

\begin{figure}[ht]
\begin{center}
   \includegraphics[width=1.0\linewidth]{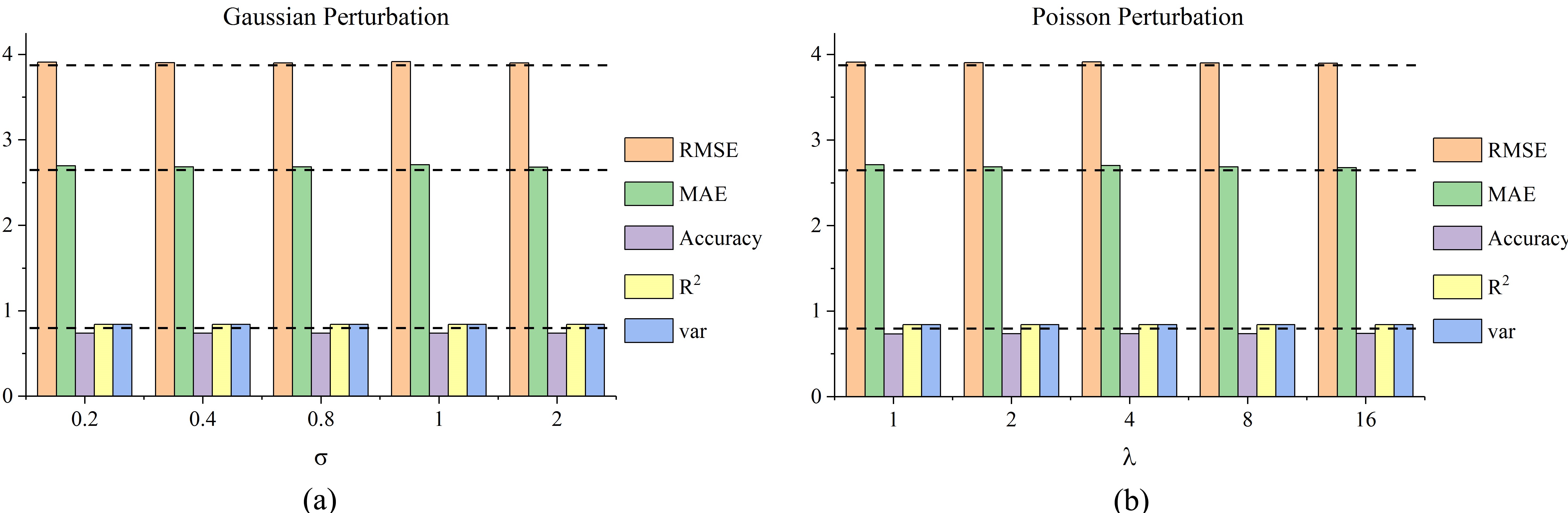}
   \end{center}
   \caption{ SZ-taxi: perturbation analysis.}
\label{fig:8}
\end{figure}

\begin{figure}[ht]
\begin{center}
   \includegraphics[width=1.0\linewidth]{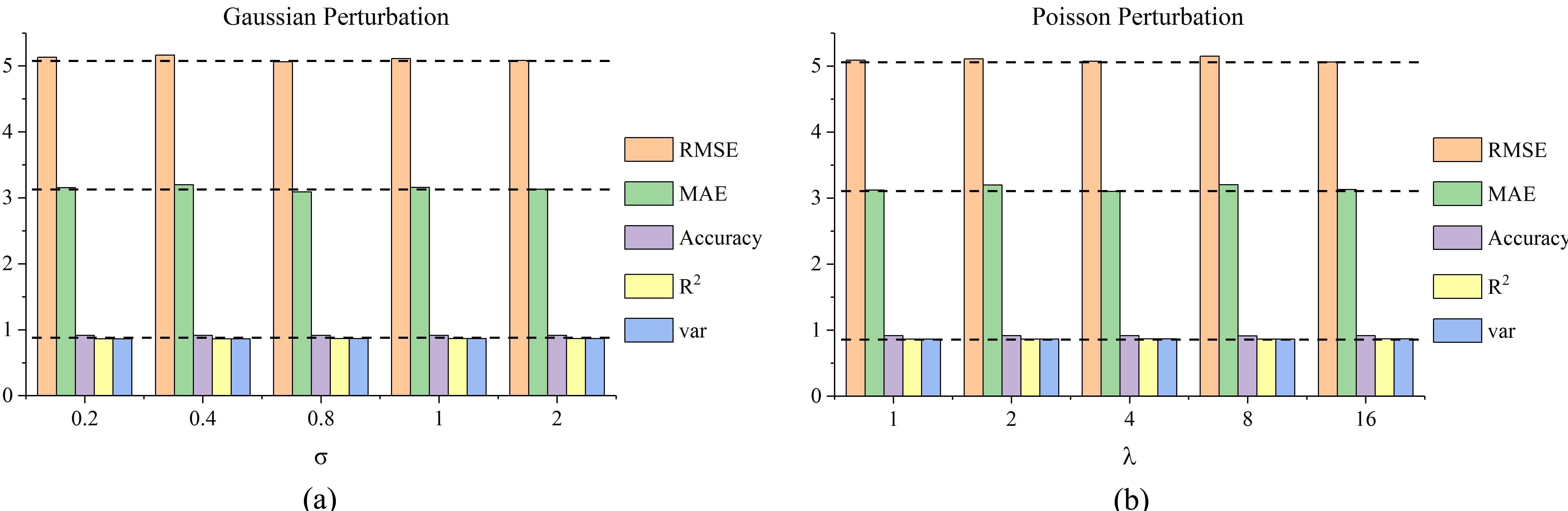}
   \end{center}
   \caption{Los-loop: perturbation analysis.}
\label{fig:9}
\end{figure}

\par The experimental results based on SZ\_taxi are shown in Fig. \ref{fig:8}. The results of adding Gaussian noise are shown in Fig. \ref{fig:8}(a), where the x- and y-axes show the changes in $\sigma$ and in different evaluation metrics, respectively. Different colors represent various metrics. Similarly, the results of adding Poisson noise are shown in Fig. \ref{fig:8}(b). The values of different evaluation metrics remain basically the same regardless of the changes in $\sigma/\lambda$. Hence, the proposed model can remarkably resist noise and process strong noise problems. 

\par The experimental results based on Los\_loop  are consistent with experimental results based on SZ\_taxi (Fig. \ref{fig:9}). Therefore, the A3T-GCN model can remarkably resist noise and still obtain stable forecasting results under Gaussian and Poisson perturbations.

\begin{figure}
\centering
\subfigure[15 minutes]{
\begin{minipage}[b]{1.0\linewidth}
\includegraphics[width=1.0\linewidth]{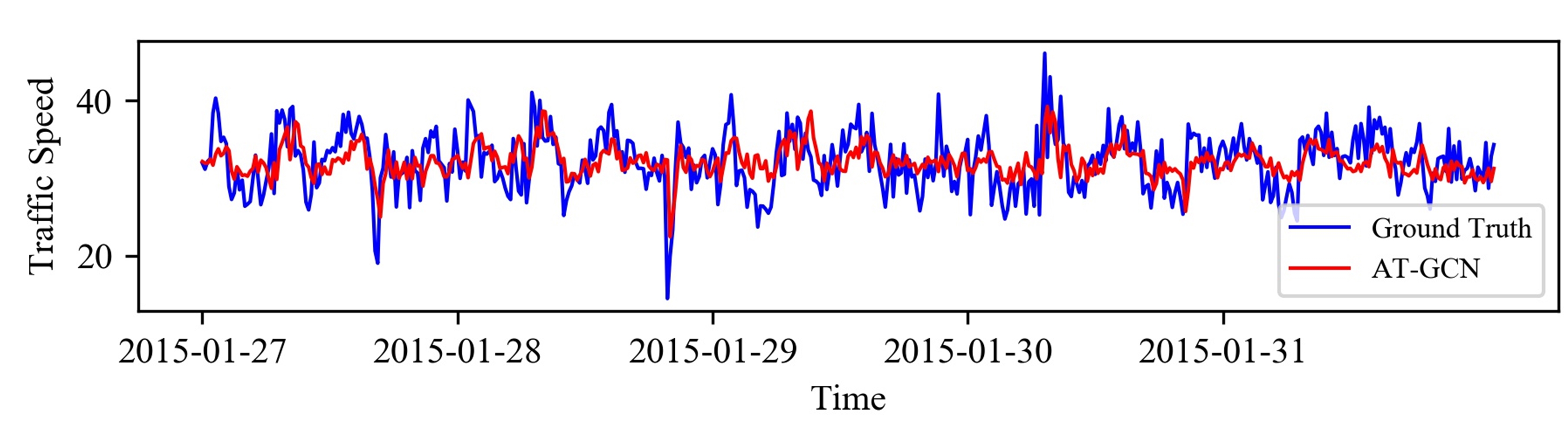} \\
\end{minipage}
}
\subfigure[30 minutes]{
\begin{minipage}[b]{1.0\linewidth}
\includegraphics[width=1.0\linewidth]{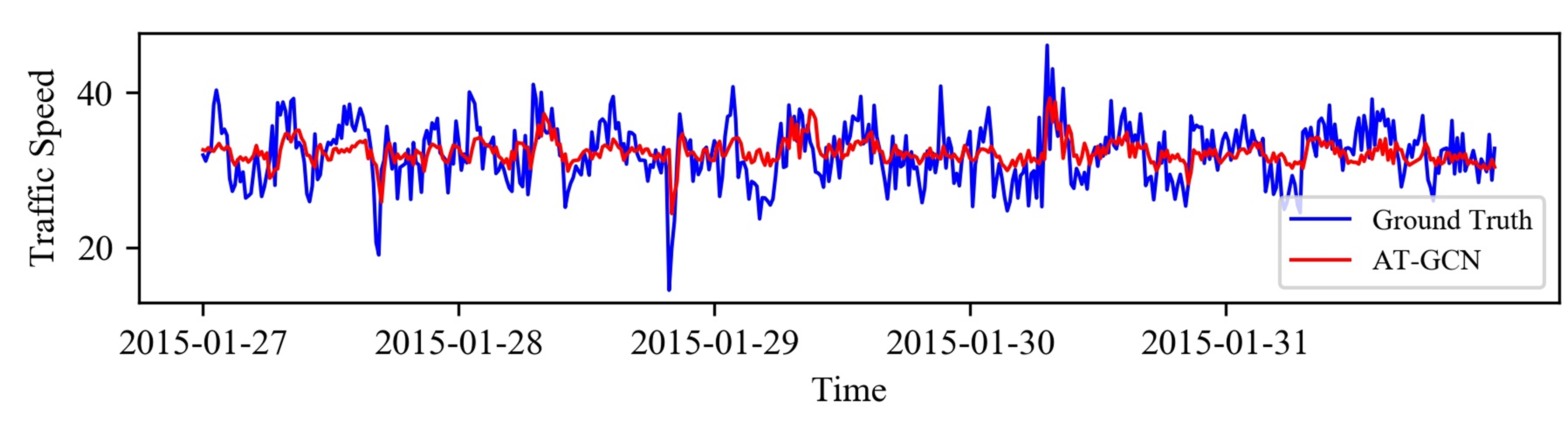} \\
\end{minipage}
}
\subfigure[45 minutes]{
\begin{minipage}[b]{1.0\linewidth}
\includegraphics[width=1.0\linewidth]{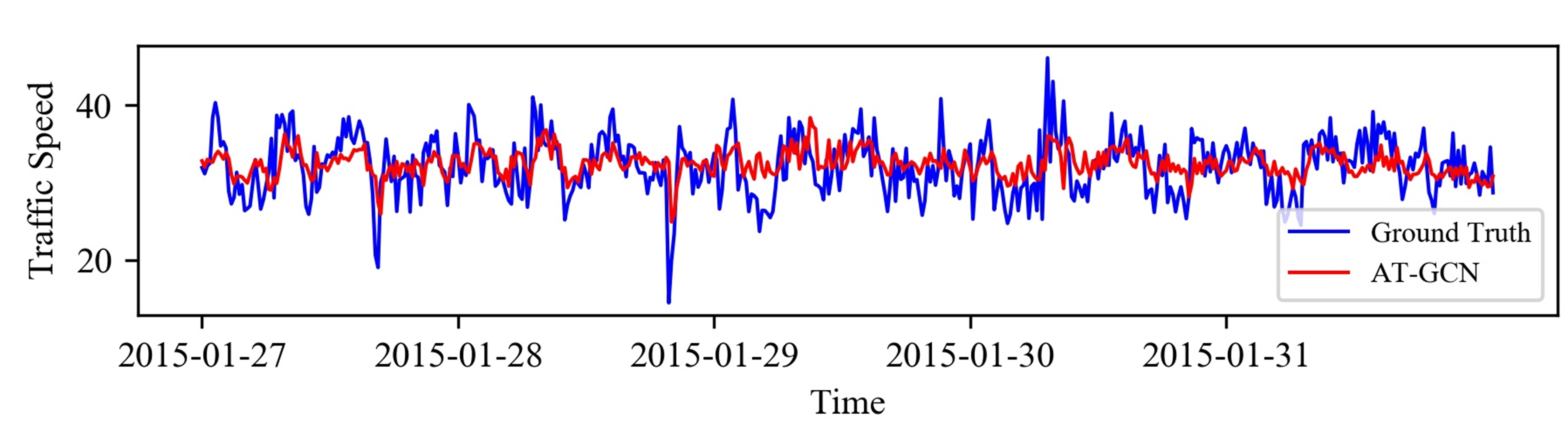} \\
\end{minipage}
}
\subfigure[60 minutes]{
\begin{minipage}[b]{1.0\linewidth}
\includegraphics[width=1.0\linewidth]{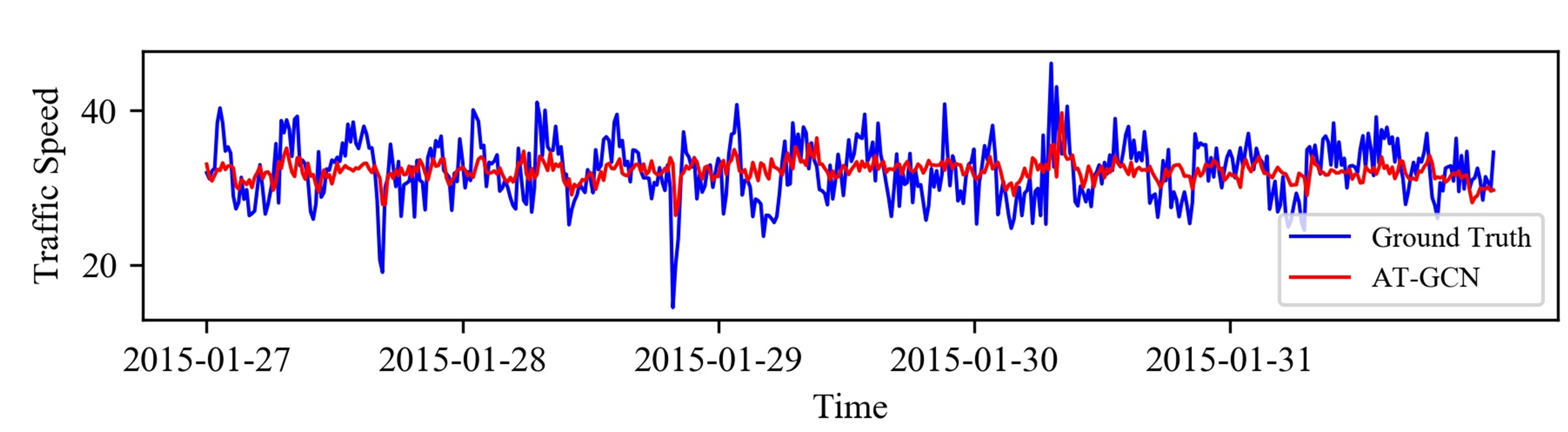} \\
\end{minipage}
}

\caption{The visualization results for prediction horizon of 15, 30, 45, 60 minutes (SZ-taxi).}
\label{fig:10}
\end{figure}

\begin{figure}
\centering
\subfigure[15 minutes]{
\begin{minipage}[b]{1.0\linewidth}
\includegraphics[width=1.0\linewidth]{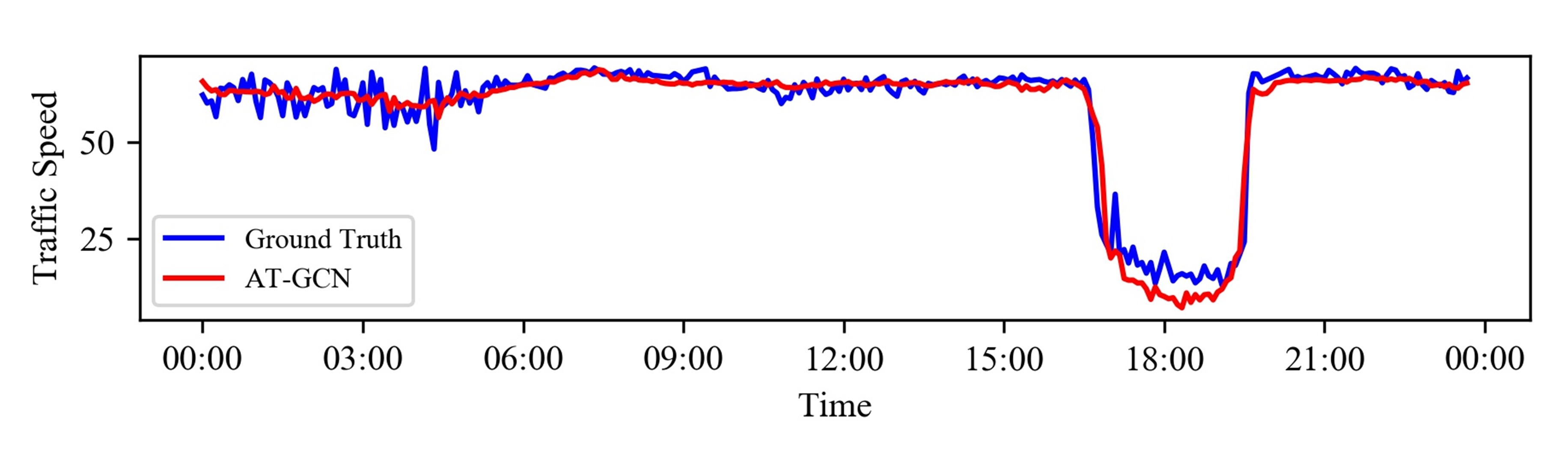} \\
\end{minipage}
}
\subfigure[30 minutes]{
\begin{minipage}[b]{1.0\linewidth}
\includegraphics[width=1.0\linewidth]{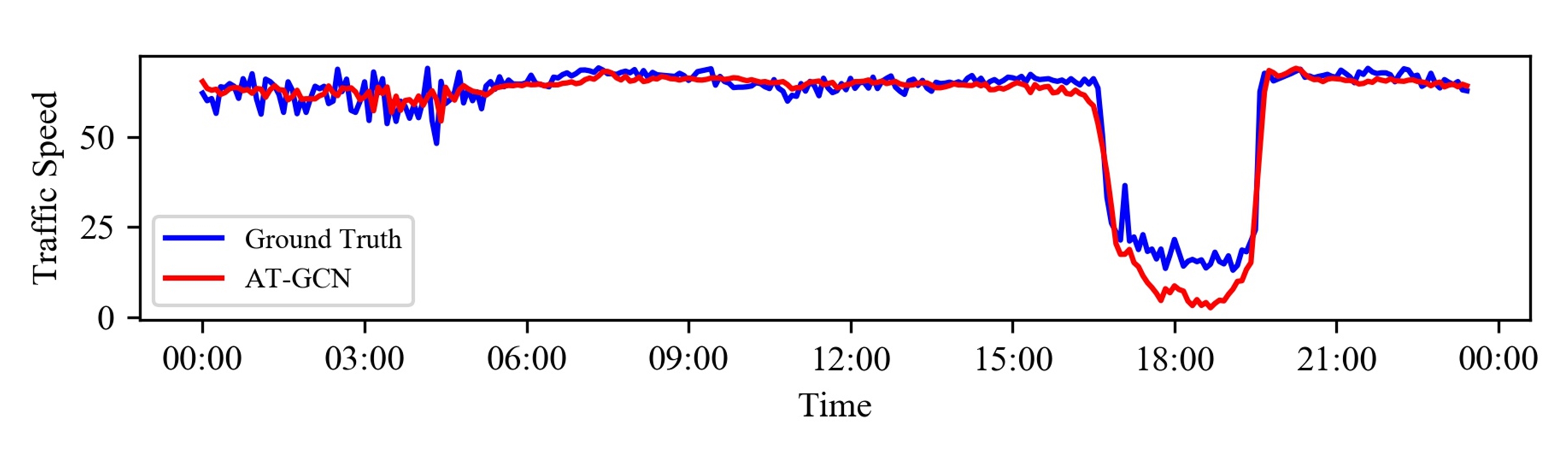} \\
\end{minipage}
}
\subfigure[45 minutes]{
\begin{minipage}[b]{1.0\linewidth}
\includegraphics[width=1.0\linewidth]{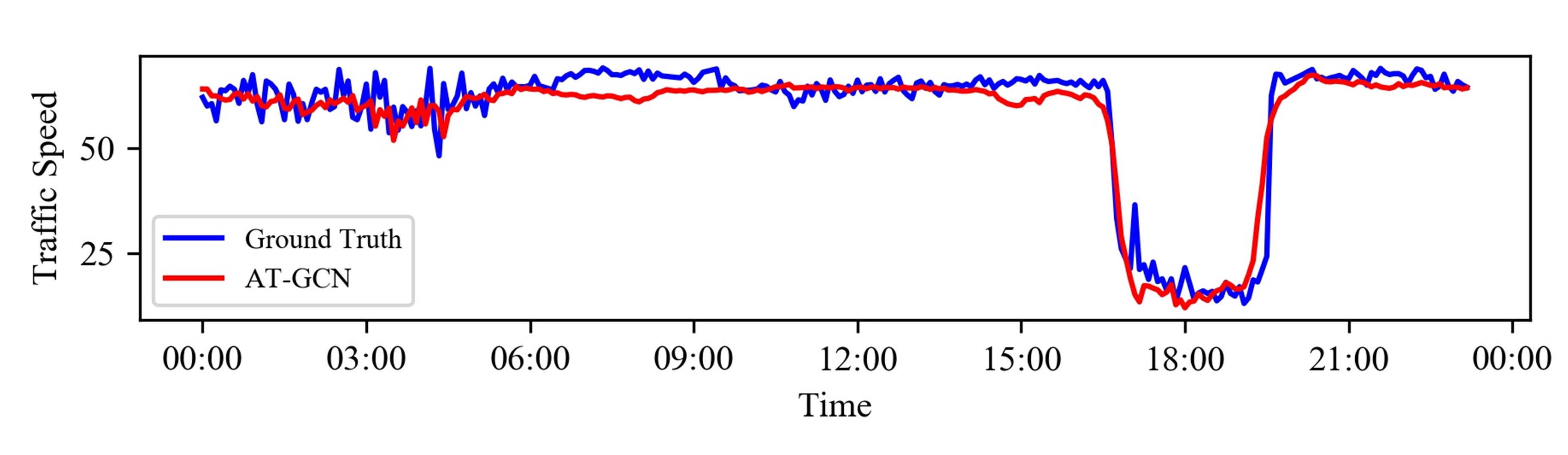} \\
\end{minipage}
}
\subfigure[60 minutes]{
\begin{minipage}[b]{1.0\linewidth}
\includegraphics[width=1.0\linewidth]{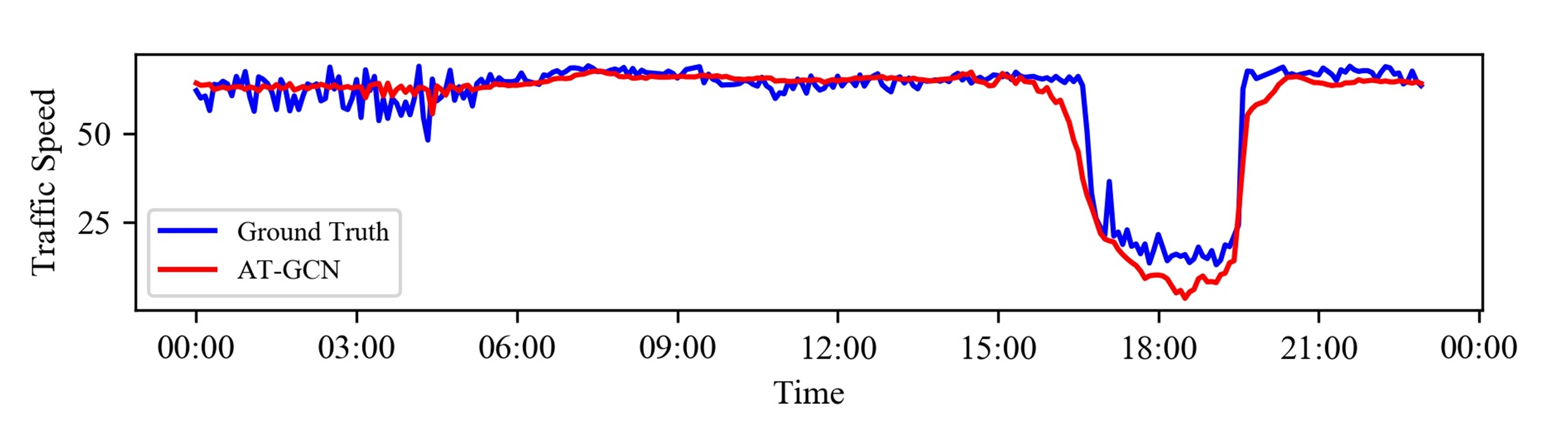} \\
\end{minipage}
}

\caption{The visualization results for prediction horizon of 15, 30, 45, 60 minutes (Los-loop).}
\label{fig:11}
\end{figure}

\subsection{Visualized analysis}
\par The forecasting results of A3T-GCN model based on two real datasets are visualized for a good explanation of the model.
\par (1) SZ-taxi: We visualize the result of one road on January 27, 2015. Visualization results in 15, 30, 45, and 60 minutes of time series are shown in Fig. \ref{fig:10}.
\par (2) Los-loop: Similarly, we visualize one loop detector data in Los-loop dataset. Visualization results in 15, 30, 45, and 60 minutes are shown in Fig. \ref{fig:11}.
\par In sum, the predicted traffic speed shows similar variation trend with actual traffic speed under different time series lengths, which suggest that the A3T-GCN model is competent in the traffic forecasting task. This model can also capture the variation trends of traffic speed and recognize the start and end points of rush hours. The A3T-GCN model forecasts traffic jam accurately, thereby proving its validity in real-time traffic forecasting.

\section{Conclusions}
\par A traffic forecasting method called A3T-GCN is proposed to capture global temporal dynamics and spatial correlations simultaneously and facilitates traffic forecasting. The urban road network is constructed into a graph, and the traffic speed on roads is described as attributes of nodes on the graph. In the proposed method, the spatial dependencies are captured by GCN based on the topological characteristics of the road network. Meanwhile, the dynamic variation of the sequential historical traffic speeds is captured by GRU. Moreover, the global temporal variation trend is captured and assembled by the attention mechanism. Finally, the proposed A3T-GCN model is tested in the urban road network-based traffic forecasting task using two real datasets, namely, SZ-taxi and Los-loop. The results show that the A3T-GCN model is superior to HA, ARIMA, SVR, GCN, GRU, and T-GCN in terms of prediction precision under different lengths of prediction horizon, thereby proving its validity in real-time traffic forecasting. 


%



\ifCLASSOPTIONcompsoc
  \section*{Acknowledgments}
\else
  \section*{Acknowledgment}
\fi

This work was supported by the National Science Foundation of China [grant numbers 41571397, 41501442, 41871364, 51678077 and 41771492].

\ifCLASSOPTIONcaptionsoff
  \newpage
\fi



%

{\small
\bibliographystyle{plain}
\bibliography{A3T-GCN}
}

\end{document}